\documentclass[letterpaper, 10 pt, conference]{ieeeconf} 

\IEEEoverridecommandlockouts                              
\usepackage{graphics} 
\usepackage{epsfig} 
\usepackage{mathptmx} 
\usepackage{times} 
\usepackage{amsmath} 
\usepackage{amssymb}  

\usepackage{graphicx}
\usepackage{subcaption}
\usepackage{algorithm}
\usepackage{algorithmic}
\usepackage{tikz}
\usepackage{placeins}
\usetikzlibrary{arrows.meta}
\usetikzlibrary{calc}
\usetikzlibrary{positioning}
\usepackage{environ}
\usepackage{multicol}
\let\labelindent\relax
\usepackage{enumitem} 
\usepackage{booktabs}
\usepackage{tikz}
\usepackage{commath}
\usepackage{url}
\usepackage{pifont}
\newcommand{\cmark}{\ding{51}}%
\newcommand{\xmark}{Fail}%

\newtheorem{theorem}{Theorem}
\newtheorem{definition}{Definition}

\newtheorem{proposition}{Proposition}

\title{Adversarial Training is Not Ready for Robot Learning}

\author{Mathias Lechner$^{1}$, Ramin Hasani${^2}$, Radu Grosu${^3}$, Daniela Rus$^{2}$, Thomas A. Henzinger${^1}$
\thanks{$^{1}$Institute of Science and Technology Austria (IST Austria)}%
\thanks{$^{3}$Massachusetts Institute of Technology (MIT), Cambridge, MA, USA}%
\thanks{$^{2}$Technische Universit\"{a}t Wien (TU Wien), 1040 Vienna, Austria}%
}

\begin{document}

\maketitle
\thispagestyle{empty}
\pagestyle{empty}


\begin{abstract}
Adversarial training is an effective method to train deep learning models that are resilient to norm-bounded perturbations, with the cost of nominal performance drop. 
While adversarial training appears to enhance the robustness and safety of a deep model deployed in open-world decision-critical applications, counterintuitively, it induces undesired behaviors in robot learning settings. 
In this paper, we show theoretically and experimentally that neural controllers obtained via adversarial training are subjected to three types of defects, namely transient, systematic, and conditional errors. 
We first generalize adversarial training to a safety-domain optimization scheme allowing for more generic specifications. We then prove that such a learning process tends to cause certain error profiles. 
We support our theoretical results by a thorough experimental safety analysis in a robot-learning task. Our results suggest that adversarial training is not yet ready for robot learning. 
\end{abstract}


\section{Introduction}
In this work, we discover that improving the safety of vision-based robot learning systems by adversarial training results in undesired side-effects of the robot's real-world behavior. Training deep neural networks while accounting for adversarial examples robustifies the model to these visually imperceptible perturbations. This process trades nominal performance gained by standard empirical risk minimization (ERM) learning techniques, with worst-case performance under norm-bounded input perturbations \cite{kurakin2016adversarial,madry2017towards,xie2019feature}.

Adversarial training has been mainly studied in image classification settings \cite{biggio2013evasion,szegedy2013intriguing,goodfellow2014explaining,carlini2017adversarial,stutz2019disentangling,salman2020adversarially}, which exclusively focused on how much adversarial training trades nominal for robust test accuracy.
While these metrics resemble the performance in static image classification tasks, robotic control tasks are inherently continuous and highly dynamic. 
Consequently, pure accuracy might not reflect the underlying performance of a robotic system accurately.
For instance, for a closed control-loop, stability might be the highest priority, whereas faithfulness could be of high priority for vehicle routing algorithms. 

In this work, we study how the nominal performance drop introduced by adversarial training methods is distributed over the real-world behavior in vision-based robot learning tasks.
First, we propose safety-domain training, a generalization of adversarial training, which allows us to incorporate more general forms of safety specifications as secondary training objectives.
We introduce a theoretical framework for characterizing error behaviors of learned controllers for robotic tasks. We then prove how safety-domain training changes the learned agent's error-profile depending on the enforced safety specification.
We test our training algorithm and confirm the consequences of our theory on an experimental case-study of an autonomous carrier robot with a variety of safety and robustness specifications.
Based on our observations, we claim that:

\begin{figure}
    \centering
    \includegraphics[width=0.9\linewidth]{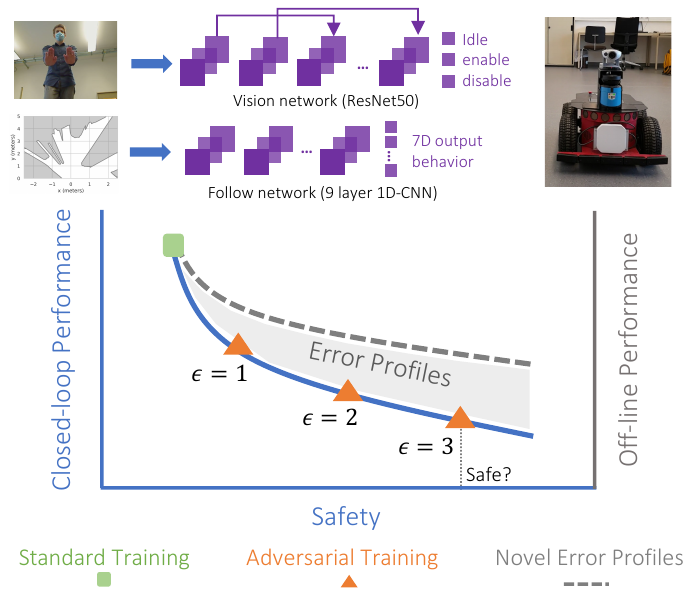}
    \caption{Adversarial training gives rise to new error profiles that significantly decrease a neural controller's performance.}
    \label{fig:intro_figure}
    \vspace{-5mm}
\end{figure}

\textit{Adversarial training is not yet ready for robot learning.}

\noindent More precisely, our experiments demonstrate that models trained by standard empirical risk minimization yield the best robotic performance in real-world scenarios.
Counterintuitively, the best-performing agents are also vulnerable to making the robot crash under adversarial patterns. 
Conversely, while the models learned via our safety-domain training are provable immune to such worst-case behavior, they significantly perform worse in real-world scenarios.
In particular, we observed that the strictness of the specifications enforced using adversarial training is the most dominant factor in determining the expected real-world robotic performance.
Our empirical evaluations confirm that adversarial training of neural controllers requires rethinking before reliably using them in robot learning schemes \cite{lechner2020gershgorin,brunnbauer2021model}. 

\noindent\textbf{Summary of Contributions.} 
\begin{enumerate}
\setlength\itemsep{0em}
    \item Formulate a generalization of adversarial training algorithms called safety-domain training with guarantees.
    \item Theoretical framework consisting of erroneous behavior characterizations and induced error-profile by our safety-domain training algorithm.
    \item Experimental confirmation of our framework on an image- and LiDAR-based autonomous carrier robot in various environments and assessment scenarios. 
\end{enumerate}

\section{Related Works}
\noindent\textbf{Adversarial Training.}
Adversarial training has led to significant improvements of deep models' resiliency to imperceptible perturbations. This was shown both empirically \cite{madry2017towards,miyato2018virtual,balaji2019instance,zhang2019theoretically} and with certification \cite{lecuyer2019certified,weng2018towards,wong2018provable,raghunathan2018certified,cohen2019certified,salman2019provably,yang2020randomized}. An emerging line of work suggests that the representations learned by adversarially trained models resemble visual features as perceived by humans more accurately compared to standard networks \cite{ilyas2019adversarial,engstrom2019learning,santurkar2019image,allen2020feature,kim2019bridging,kaur2019perceptually}. In contrast, a large body of work tried to characterize the trade-off between a model's robustness and accuracy when trained by adversarial learning schemes \cite{tsipras2018robustness,bubeck2019adversarial,su2018robustness,raghunathan2019adversarial}. Some gradient issues, such as gradient obfuscation \cite{athalye2018obfuscated,uesato2018adversarial}, during training, seemed to play a role in the mediocre performance of the models. Nevertheless, adversarially trained networks also showed to maintain their robustness properties \cite{shafahi2019adversarially} as well as their accuracy \cite{utrera2020adversarially,salman2020adversarially} in transfer learning settings. 

This work shows that despite the vast success of adversarially trained models in obtaining robustness properties on vision-based classification tasks, they can introduce novel error profiles in robot learning schemes. Our work aims to identify and report these profiles to enable the practical use of adversarial training in safety-critical applications.

\noindent\textbf{Adversarial Training for Safe Robot Learning.} Related approaches can be grouped into three categories; i) adversarial learning as a data augmentation technique; ii) Hand-crafted perturbation distribution; and iii) Task-specific models.
\begin{enumerate}[label=(\roman*)]
    \item \textbf{Adversarial learning as a data augmentation technique --} A couple of recent works characterized generative adversarial networks (GANs) \cite{goodfellow2014generative} as a data augmentation method to enhance neural controllers' transferability. For example, \cite{chen2020adversarial} used GAN-based training for robotic visuomotor control, and \cite{porav2018adversarial} explored GANs to determine robust metric localization by using appearance transfer (e.g., day to night transformation of input images). These methods fundamentally differ from safety-related adversarial training frameworks \cite{szegedy2013intriguing} that we explore in this paper, as they refer to methods useful for data augmentation.
    \item \textbf{Hand-crafted perturbation distribution --} invariant sets, i.e., hand-crafted changes in underlying data distribution such as change of a gripper's appearance and objects' color used in task-relevant adversarial imitation learning
\cite{zolna2019task}. These approaches require a simulator capable of generating domain-specific perturbations and are mainly designed for training performant agents. Our paper discusses a more general setting where we do not require domain-specific attributes.
    \item \textbf{Task-specific models --} Adversarial training in task-specific domains such as motion planning \cite{janson2018safe,shi2020robust,innes2020elaborating} and localization \cite{yang2020map} has been used for enhancing robustness. Moreover, in reinforcement learning (RL) environments, adversarial training benefited agents in competitive scenarios such as active perception \cite{shen2019active}, interaction-aware multi-agent tracking, and behavior prediction \cite{li2019interaction} and identifying weaknesses of a learned policy \cite{pan2019risk,kuutti2020training}. 
    These works do not evaluate existing general methods but propose tailored solutions for the specific task under-test. Our paper focuses on the broad vision-based robot learning problems that use contemporary adversarial training for enhancing safety. 
\end{enumerate}

\section{Error Profiles in Robot learning}
Training a neural network $f_\theta$ in supervised learning, considers finding the best approximation function $f_\theta: x \mapsto y$ parameterized by $\theta$, that maps the input data $x$ to labels $y$. In a robotic learning setting, the data $(x,y)$ is taken from a data distribution over a finite subset $\mathbb{D}$ of the functional relation $\mathbb{R}^{N_x}\times \mathbb{R}^{N_y}$, i.e., sensor values and motor commands have finite precision. 
Limited training data, noise in the learning process \cite{hasani2019machine}, and inadequate causal modeling \cite{scholkopf2019causality,hasani2020liquid,lechner2020neural,lechner2020learning} prevent the network from achieving a perfect mapping of the ground truth dependency between $x$ and $y$. Consequently, these imperfections lead to errors during test time.
In robot learning settings, we characterize these errors made by a neural controller by three categories: Systematic errors, transient errors, and conditional errors. Our objective is to show that these error profiles occur and lead to mediocre performance even when the neural controller is trained by safety-domain training methods such as adversarial training \cite{szegedy2013intriguing}. Let us first formally define these error profiles:

\begin{definition}[Transient error]\label{def:transient}
Given a loss function $\mathcal{L}$, a neural network $f_\theta$, a threshold $\eta>0$ and neighborhood $\varepsilon>0$, we call a point $(x',y')$ transient error if 
\begin{equation}
\mathcal{L}(y',f_\theta(x')) > \eta \text{ and }  \mathcal{L}(\tilde{y},f_\theta(\tilde{x})) < \eta,
\end{equation}
for all $(\tilde{x},\tilde{y})$ where $0 < ||\tilde{x}-x'||\leq \varepsilon$.
\end{definition}

\begin{definition}[Systematic error]\label{def:systematic}
Given a loss function $\mathcal{L}$, a neural network $f_\theta$, a threshold $\eta>0$, we say $f_\theta$ suffers from a systematic error if 
\begin{equation}
\mathcal{L}(y,f_\theta(x)) > \eta,
\end{equation}
for all $(x,y) \in \mathbb{D}$.
\end{definition}

\begin{definition}[Conditional error]
Given a loss function $\mathcal{L}$, a neural network $f_\theta$ and a threshold $\eta>0$, then we call a domain $\mathcal{D} = \{(x,y)|(x,y) \in \mathbb{D}\}$ a conditional error if 
\begin{equation}
\underset{(x,y) \in \mathcal{D}^{}}{\mathbb{E}}\big[\mathcal{L}(y,f_\theta(x)\big] > \eta ,
\end{equation}
and
\begin{equation}
\underset{(x,y) \in \mathcal{D}^{c}}{\mathbb{E}} \big[ \mathcal{L}(y,f_\theta(x)\big] < \eta.
\end{equation}
\end{definition}
\begin{figure}
\centering
\includegraphics[width=0.4\textwidth]{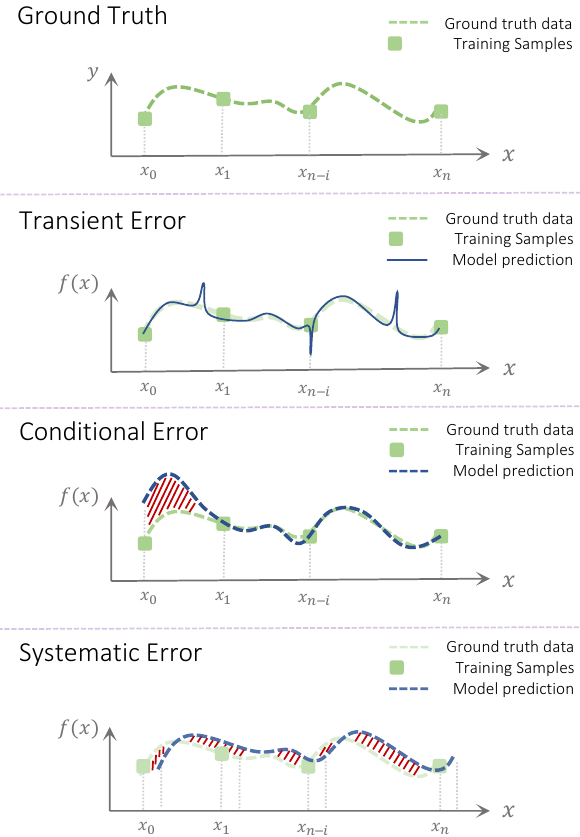}
\caption{Different types of errors that can occur when fitting a network $f(x) =y$}
\label{fig:errormodes}
\end{figure}

Fig. \ref{fig:errormodes} illustrates these error types schematically for a single-dimensional learning problem. Transient errors might induce divergent behavior at the evaluation time (e.g., see the model's approximation at the evaluation step $x_{n-i}$) in the second chart). Conditional errors can result in local mismatches that affect an agent's local behavior to reach unsafe states. Systematic errors lead to distribution and base-line shifts for the entirety of the sample data. Next, we define a safety-domain robot learning scheme equipped with adversarial training and explore if this framework can give rise to the error types identified above.

\section{Learning with worst-cases in mind}
This section defines a generalization of adversarial training by relaxing the $ \epsilon $-neighborhood for arbitrary domains. We call this approach \textit{Safety-Domain Training}. We then explain how to solve the inner optimization loop of safety-domain training, either by empirical or certified safety methods and illustrate the resulting method in Algorithm \ref{algorithm:safety}. 

Standard training of a neural network $f_\theta$ concerns optimizing $\theta$ to minimize the empirical risk as follows:
\begin{equation}\label{eq:risk}
     \min_{\theta} \frac{1}{n} \sum_{i=1}^{n}  \mathcal{L}\big(y_i, f_\theta(x_i)\big),
\end{equation}
where $\mathcal{L}$ is a loss function and $\{(x_i,y_i)|i=1,2,\dots n\}$ the training samples. The optimization performed by mini-batch stochastic gradient descent (SGD) yields the best performing networks.

Adversarial training extends the objective in Eq. \ref{eq:risk} by replacing each input $x_i$ by an adversarial input $\tilde{x}_i$ within the $\varepsilon$-neighborhood of $x_i$. Formally:

\begin{definition}[Adversarial training \cite{szegedy2013intriguing}]
Let $f_\theta$ be a neural network, $\{(x_i,y_i)|i=1,2,\dots n\}$ the training samples, $\mathcal{L}$ the loss function, and $\varepsilon > 0$ the adversarial attack radius. Then adversarial training optimizes the criterion
\begin{equation}\label{eq:defadv}
    \min_{\theta} \frac{1}{n} \sum_{i=1}^{n} \max_{\Tilde{x}: ||x_i-\tilde{x}||<\varepsilon} \mathcal{L}\big(y_i, f_\theta(\Tilde{x})\big) 
\end{equation}
\end{definition}
We generalize adversarial training to a more generic safety-domain training. In particular, we replace the $\varepsilon$-neighborhoods of the training samples by arbitrary domains, i.e., labelled and connected sets.

\vspace{+1.5mm}
\begin{definition}[Safety-domain training]
Let $f_\theta$ be a neural network, $\{(x_i,y_i)|i=1,2,\dots n\}$ the training samples, $\mathcal{L}$ the loss function, and $\{(D_i,z_i)|i=1,2,\dots k\}$ the safety domains. Then safety-domain training optimizes the criterion
\begin{equation}\label{eq:defsafety}
    \min_{\theta}  \big[ \frac{1}{n} \sum_{i=1}^{n} \mathcal{L}\big(y_i, f_\theta(x_i)\big) +\lambda   \frac{1}{k} \sum_{i=1}^{k} \max_{\Tilde{x} \in D_i} \mathcal{L}\big(z_i,f_\theta(\Tilde{x})\big)  \big],
\end{equation}
where the hyperparameter $\lambda$ specifies the tradeoff between optimizing the empirical training risk and the worst-case risk on the safety-domains.
\vspace{+1.5mm}
\end{definition}

Safety-domain training generalizes adversarial training and can be shown intuitively as follows:

\begin{proposition}
\label{cor:adv}
[Safety-domain training is a generalization of adversarial training]
The criterion Eq \eqref{eq:defsafety} is a generalization of the adversarial training objective in Eq \eqref{eq:defadv}. In particular, this is true as we define $D_i:=\{\tilde{x}: ||\tilde{x}-x_i||<\varepsilon \}$ with $z_i:=y_i$ and the training samples $\{\}$.
\end{proposition}

\begin{algorithm}[b]
\caption{Safety-domain training with guarantees}
\label{algorithm:safety}
\begin{algorithmic}
\STATE \textbf{Input:} Training data $\{(x_i,y_i)|{i=1\dots n}\}$, Safety domains $\{(z_i,D_i)|{i=1\dots k}\}$
\STATE \textbf{Parameters:}  safety threshold $\delta$, batch sizes $b_t,b_s$
\STATE \ \ \ \ \ \ \ Learning rate $\alpha$, minimum training epochs $i_{\text{min}}$.
\STATE $\text{safety\_bound} = \infty$ 
\WHILE{$i < i_{\text{min}}$ \textbf{and} $\text{safety\_bound} > \delta$} 
\STATE i = i+1
\STATE $(\tilde{x},\tilde{y}) = \text{sample\_batch}(b_t,\{(x_i,y_i)|{i=1\dots n}\})$
\STATE $(\tilde{z},\tilde{D}) = \text{sample\_batch}(b_s,\{(z_i,D_i)|{i=1\dots k}\})$
\STATE $\nabla = \frac{\partial }{\partial \theta} \frac{1}{b_t}\sum_{i=1}^{b_t}\mathcal{L}\big(\tilde{y}_i,f_\theta(\tilde{x}_i\big)$
\STATE $\nabla = \nabla + \frac{\partial }{\partial \theta} \lambda   \frac{1}{b_s} \sum_{i=1}^{b_s} \max_{x \in \tilde{D}_i} \mathcal{L}\big(\tilde{z}_i,f_\theta(x)\big)$
\STATE $\theta = \theta - \alpha \nabla$
\STATE $\text{safety\_bound} = \max_{i=1\dots k} \max_{x \in D_i} \mathcal{L}\big(z_i,f_\theta(x)\big) $
\ENDWHILE
\RETURN $\theta$
\end{algorithmic}
\end{algorithm}

\textbf{Empirical vs. certified safety}
In practice, we have two options on how we solve the inner maximization step of safety-domain training. The first option is to perform several steps of projected gradient descent \cite{madry2017towards}. While this approach is computationally efficient and straightforward to implement, it provides no true worst-case guarantees as SGD does not ensure convergence to the global optimum. In practice, empirical approaches are often used for adversarial training of classifiers to account for computational complexity. 

A more rigorous approach, albeit expensive, is to compute an upper-bound of the loss of each safety domain and minimize the upper bound via stochastic gradient descent \cite{lecuyer2019certified,gruenbacher2020verification}. While computing an upper-bound of a network's output is difficult and may overestimate the true maximum, it provides certified guarantees on the worst-case loss.
The interval bound propagation method falls into the category in which the upper bound of the loss is computed by an interval arithmetic abstraction of the network \cite{gowal2018effectiveness,huang2019achieving}. 
The main difficulty of such certified approaches is to scale the training to large networks.

Algorithm \ref{algorithm:safety} represents our framework for training an adversarially robust neural controller with safety guarantees. 

\section{Adversarial Training Not Ready for Robot Learning}
In this section, we show theoretically that adversarial training and even its generalization result in unexplored error profiles that lead to unsafe behavior. To construct our theory, we first describe a set of required assumptions.

\subsection{Assumptions}
To build the theory, we should rule out ill-posed edge-cases, thus relying on the following assumptions:
1) \textit{Bounded generalization} The training loss provides a lower bound on the generalized loss \cite{bousquet04}, i.e., the expected loss over the data distribution.
2) \textit{Bounded sets} Safety domains are bounded subsets of the data domain $\mathbb{D}$.
3) \textit{Non-conflicting data} Safety-domains and ground truth data are non-conflicting, i,e., for every sample $(x_i,y_i)$ and safety domain $(D_j,z_j)$ it holds that if $x_i \in D_j \implies y_i=z_j$

\begin{theorem}
	\label{thm:safety}
	[Safety-domain training tend to cause conditional errors]
	Let $f$ be a neural network, $\mathcal{L}$ be a loss function, $\{(x_i,y_i)|i=1,2,\dots m\}$ be the iid training data, and $\theta$ be the weights obtained by Algorithm \ref{algorithm:safety} with the safety-domains $\{(D_i,z_i)|i=1,2,\dots k\}$ and the threshold $\delta$. Moreover, we assume $\delta$ to be a lower bound of the total training loss in Equation (\ref{eq:defsafety}), i.e., ruling out cases where fitting $f$ to the underlying data distribution is trivial.
	Let $(x,y)$ be an arbitrary sample from the underlying data distribution, then
	\begin{equation}
	\mathbb{E}\big[\mathcal{L}(y,f_\theta(x))\big| x \in \bigcup_{i=1}^{k} D_i  \big] \leq  \mathbb{E}\big[\mathcal{L}(y,f_\theta(x))\big| x \notin \bigcup_{i=1}^{k} D_i  \big] 
	\end{equation}
\end{theorem}
\vspace{0.5cm}
\begin{proof}
	For $x \in  D_i$ we have 
	\begin{align*}
	\mathcal{L}\big(y,f_\theta(x)\big) &\leq \max_{\Tilde{x} \in D_i} \mathcal{L}\big(z_i,f_\theta(\Tilde{x})\big)\\
	& \leq \max_j \max_{\Tilde{x} \in D_j} \mathcal{L}\big(z_j,f_\theta(\Tilde{x})\big)\\
	& = \delta.
	\end{align*}
	Thus 
	\begin{equation*}
	\mathbb{E}\big[\mathcal{L}(y,f_\theta(x))\big| x \in \bigcup_{i=1}^{k} D_i  \big] \leq \delta.
	\end{equation*}
	For  $x \notin \bigcup_{i=1}^{k} D_i$, we know from our generalization bound
	\begin{align*}
	\mathbb{E}\big[\mathcal{L}(y,f_\theta(x))\big| x \notin \bigcup_{i=1}^{k} D_i   \big] &\geq  \frac{1}{n} \sum_{i=1}^{n}  \mathcal{L}\big(y_i, f_\theta(x_i)\big)\\ 
	& \geq \delta,	\end{align*}
	which shows the claim.
\end{proof}

In the rest of this section, we discuss two important implications of Theorem \ref{thm:safety} when adding additional assumptions about the networks and training process.
In particular, we assume
1) \textit{Locality} For any two training samples $(x_1,y_1), (x_2,y_2)$ and a trained $f_\theta$ it holds that 
\begin{align*}
||x_1-x_2|| > K \implies &\\ 
\mathbb{E}\big[\mathcal{L}(y_1,f_\theta(x_1)) \big] \approx&\  \mathbb{E}\big[\mathcal{L}(y_1,f_\theta(x_1))\big| \mathcal{L}(y_2,f_\theta(x_2)) \big],
\end{align*}
i.e., the network's performance of a sample is not influenced by far apart samples.
2) \textit{Training stability} For any training sample $(x,y)$ and two trained networks  $f_{\theta_1}, f_{\theta_2}$ it holds that 
\begin{equation*}
\mathbb{E}\big[\mathcal{L}(y,f_{\theta_1}(x)) \big] \approx \mathbb{E}\big[\mathcal{L}(y,f_{\theta_2}(x)) \big],
\end{equation*}
i.e., re-training a network does not introduce errors into the network.

\noindent\textit{Implication 1: The conditional errors introduced by the safety-domain training tends to occur near the boundaries of the safety-domains}
Let $f_\theta$ be a network obtained by safety-domain training and $f_{\theta'}$(  be a network obtained my standard training. Moreover, let $(x,y)$ be a sample with $(x,y)$ with $ \mathbb{E}[\mathcal{L}(y,f_{\theta}(x)) ] >  \mathbb{E}[\mathcal{L}(y,f_{\theta'}(x)) ]$, then 
\begin{align*}
(x,y) \in \{(x,y)| (x,y) \notin D_i \text{ and } ||x-\tilde{x}||\geq K \\
\text{ for all } (\tilde{x},\tilde{y})\in D_i \text{ and } i=1,\dots k \},
\end{align*} for some $K>0$.

\noindent The claim that $(x,y) \notin D_i$ follows simply from applying Theorem \ref{thm:safety}. The second part, i.e., $||x-\tilde{x}||\geq K$, can be shown by deriving a contradiction when assuming the opposite is true. In particular, if $||x-\tilde{x}||<K$, our locality assumption implies that  $\mathbb{E}[\mathcal{L}(y,f_{\theta}(x)) ]$ is independent from any $(\tilde{x},\tilde{y})\in D_i$. Due to this independence and our second training stability assumption, we know that the expected loss at $(x,y)$ does not change when we retrain the network without any safety-domains. However, as safety-domain training with empty safety-domains corresponds to standard training, this contradicts our assumption that $ \mathbb{E}[\mathcal{L}(y,f_{\theta}(x)) ] >  \mathbb{E}[\mathcal{L}(y,f_{\theta'}(x))]$.

\begin{proposition}
	\label{cor:errs}
	[Transient and systematic errors are special cases of conditional errors]
	Let $(x',y')$ be a transient error according to definition \ref{def:transient}, then it implies a special case of a conditional error with $\mathcal{D}={(x',y')}$ and $\mathcal{D}^c={(x,y)| (x,y) \in \mathbb{D}: 0 < ||\tilde{x}-x'||=\varepsilon}\}$, i.e., a domain with only a single element and whose underlying data domain is defined locally.
	Moreover, if $f_\theta$ suffers from a systematic error as defined in \ref{def:systematic}, then it implies a special case of a conditional error with $\mathcal{D}=\mathbb{D}$, i.e., conditional error domain is the entire data domain.
\end{proposition}
Proposition \ref{cor:errs} shows that adversarial training, i.e., when the safety-domains are small and sampled across the entire data domain, can lead to transient and systematic errors.

\section{Experimental Evaluation: A Case-study}
The objective of our experimental evaluation is to I) validate our claims empirically and II) study the different error-types on a more fine-grained level than the theory allows. In particular, we will study an end-to-end robotic control problem by learning from demonstration with deep models. Besides optimizing for high accuracy, we enforce secondary robustness and safety specifications on the networks.
We impose the specification with various strictness levels to study how the specifications affect the network's error characteristics on a fine-grained scale.

\begin{figure}[t]
    \centering
    \includegraphics[width=\linewidth]{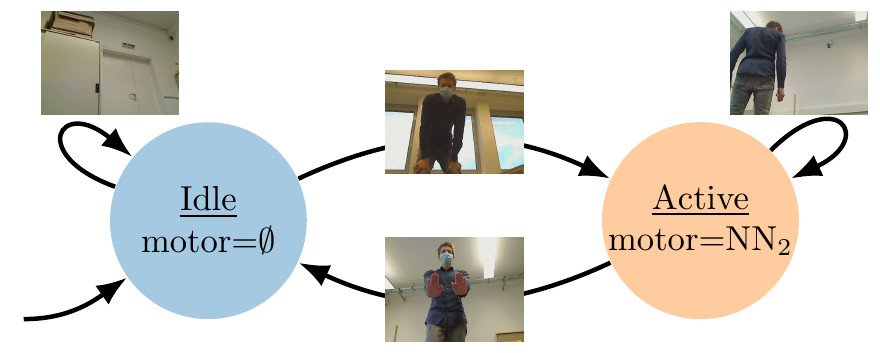}
    \caption{State machine of the high-level controller. Transitions between states are triggered by a ResNet50 image classifier. In the active state the second neural network translates the LiDAR inputs to motor outputs.}
    \label{fig:state}
    \vspace{-4mm}
\end{figure}

In our case study, we develop a controller for a vision + Lidar-based mobile robot.
A human operator enables and disables the mobile robot via visual gestures. Once activated, the robot navigates such that it always faces the human operator at a distance of roughly one meter.
The control software consists of two neural networks and a state-machine with two states. 
State transitions are triggered by a neural network (the vision network) processing the camera inputs. Fig. \ref{fig:state} shows an illustration of the state-machine and its transition profiles.
The robot's active behavior is realized via a second neural network (the follow network) that continuously translates a 2D-LiDAR scan of the environment into motor commands.
The correct behavior is entirely determined by the networks' performance, making the controller well suited for our empirical study. A video demonstration of the controller in action can be found at \url{https://youtu.be/xrgnSh1mk38}.

Our implementation approach is in contrast to traditional approaches for implementing such a robotic controller, which rely on hand-designed rules applied to infrared sensors \cite{afghani2013follow} camera inputs \cite{yoshimi2006development,munaro2012tracking}, or local localization protocols \cite{pradeep2017follow}. 
Perhaps the closest work to ours is the setup described in \cite{satake2009robust}, which uses a stereo camera setup and machine learning using Support Vector Machines (SVMs).

Our physical robot is equipped with a Sick LMS1xx 2D-LiDAR rangefinder, a Logitech RGB camera, and a 4-wheeled differential drive. Consequently, our application allows for an additional layer of safety compared to pure vision-based approaches.

\begin{figure}
    \centering
    \includegraphics[width=\linewidth]{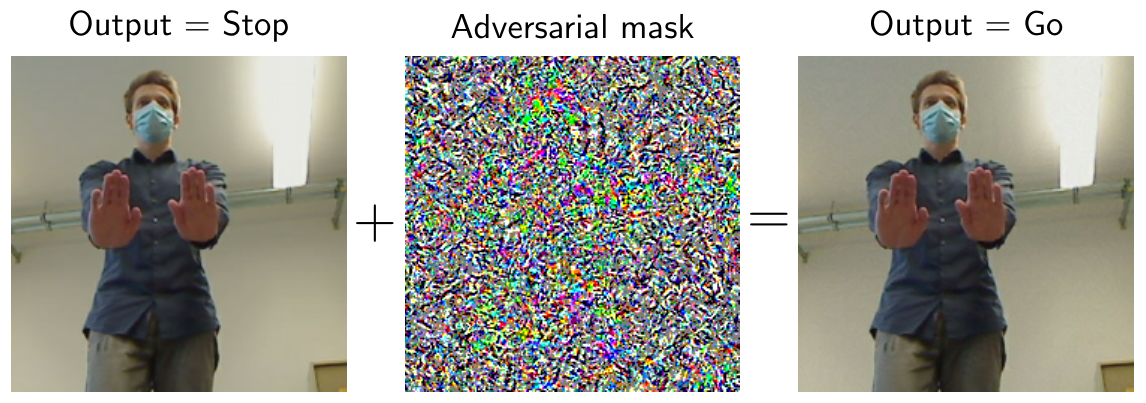}
    \caption{Visualization of an adversarial attack on our vision network. Adding an adversarial mask flips the decision from a stop command to an activation command. The adversarial mask changes each pixel by only $\pm$ 2 steps on the 8-bit color channels, making the attack imperceptible to humans. }
    \label{fig:vision_attack}
\end{figure}

For our vision network we use a ResNet50 \cite{he2016deep} pre-trained on ImageNet \cite{russakovsky2015imagenet}. The fine-tuning task concerns classifying 1825 training images in three categories, i.e., idle, enable, disable gestures, as illustrated in Fig. \ref{fig:state}. We avoid overfitting of the network by fine-tuning only the last few layers.
We train the vision network by adversarial training with the fast-gradient-sign method \cite{goodfellow2014explaining} and three different values for $\epsilon$, i.e., $l_\infty$ neighborhoods with $\varepsilon \in \{0,1,2\}$, see Fig. \ref{fig:vision_attack} for an example. Note that adversarial training with $\varepsilon=0$ is equivalent to a standard empirical risk minimization training. The training and validation accuracy is reported in Table \ref{tab:vision_acc}.

\begin{table}[b]
    \centering
    \caption{Training and validation accuracy of the vision network trained with different adversarial perturbation radii. Training accuracy represents adversarial accuracy and validation accuracy represents clean accuracy.}
    \begin{tabular}{c|c|cccc}
    \toprule
         \textbf{Level}  & \textbf{Training acc. (adversarial)} & \textbf{Validation acc. (clean)} \\
         \midrule
         0 & 99.7 \% & 98.4\%  \\
         1 & 52.0\% & 92.8\% \\
         2 & 32.5\%  & 71.9\% \\
         \bottomrule
    \end{tabular}
    \label{tab:vision_acc}
\end{table}

Our command-following network is a 9-layer 1D convolutional neural network mapping the 541-dimensional inputs to 7 possible output categories, i.e., stay, forward, left forward, right forward, backward, left backward, and right backward.
As the Follow network directly controls the motors, it potentially crashes into the human operator or an obstacle causing physical damage. To avoid such worst-case outcomes, we enforced safety-specifications on the network. In particular, we want to avoid the forward movement of the robot in case an object is in front of it. In Table \ref{tab:safety_domains}, we define four levels of safety-domains that characterize our safety requirements with increasing strictness.

\begin{table}[t]
    \centering
        \caption{Specification of the safety domains $D_i$ for the different safety levels.}
    \begin{tabular}{c|l}
    \toprule
        \textbf{Level} & \textbf{Description of safety domains $D_i$} \\
        \midrule
        0 & $D_i = \emptyset$\\\hline
        1 & $D_i=\big\{x\big| 0\leq x_j \leq 0.2 \text{ for } j\in \{i-1,i,i+1\} \text{ and }$\\
         &  $\qquad \qquad 0\leq x_j \leq 3 \text{ for }j \notin \{i-1,i,i+1\} \big\}$ \\\hline
        2 & $D_i=\big\{x\big| 0\leq x_i \leq 0.2 \text{ and }$\\
         &  $\qquad \qquad 0\leq x_j \leq 3 \text{ for }j \neq i \big\}$ \\\hline
        3 & $D_i=\big\{x\big| 0\leq x_i \leq 0.2 \text{ and }$\\
         &  $\qquad \qquad 0\leq x_j \leq 4 \text{ for }j \neq i \big\}$ \\
         \bottomrule
    \end{tabular}
    \caption*{\footnotesize \textbf{Setup Details:} For each level there are 240 domains, i.e., $i=150\dots 390$. The corresponding labels $z_i$ are defined as a any non-forward moving category, i.e., $z_i \in \{\text{stay}, \text{backward}, \text{left backward}, \text{right backward}\}$. The domains with increase safety level represent super-set of the lower safety level, e.g. the conditions considered at level 1 are a strict subset of the level 2 safety.  Level 1 safety only considers cases where at least three consecutive LiDAR rays are less than 20 cm, whereas one ray is enough for level 2 and 3. Level 3 differs from level 2 in terms of the upper bounds on the other rays.}
    \label{tab:safety_domains}
\end{table}
\begin{table}[b]
    \centering
    \caption{Training and validation accuracy of the follow network trained when enforcing different safety-levels.}
    \begin{tabular}{c|c|cccc}
    \toprule
         \textbf{Level}  & \textbf{Training accuracy} & \textbf{Validation accuracy}  \\
         \midrule
         0 &  98.8\% & 84.7\%  \\
         1 & 99.7\% & 76.8\% \\
         2 & 97.1\% & 73.4\% \\
         3 & 57.3\% & 53.2\%\\
         \bottomrule
    \end{tabular}
    \label{tab:follow_acc}
\end{table}
\begin{figure}[t]
    \centering
    \includegraphics[width=\linewidth]{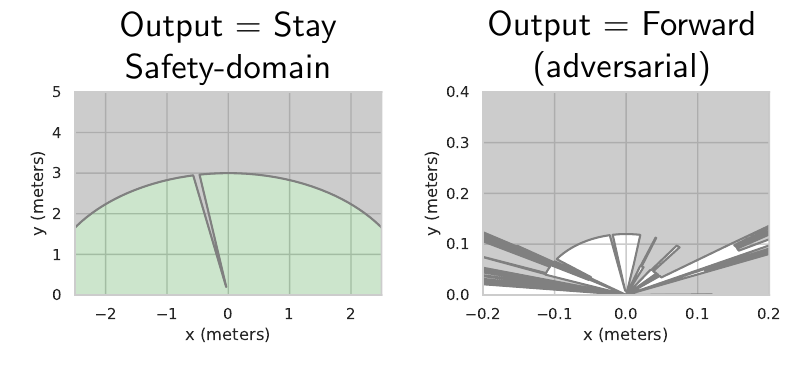}
    \caption{\textbf{Left:} Visualization of a safety-domain. No LiDAR signal in green area should be classified as a ''forward'' decision. \textbf{Right: } The network trained with standard ERM can be attacked to output a ''forward'' despite the LiDAR signal indicating a large object 10cm in front of the robot.}
    \label{fig:attacks}
    \vspace{-4mm}
\end{figure}

\begin{figure}
    \centering
    \includegraphics[width=\linewidth]{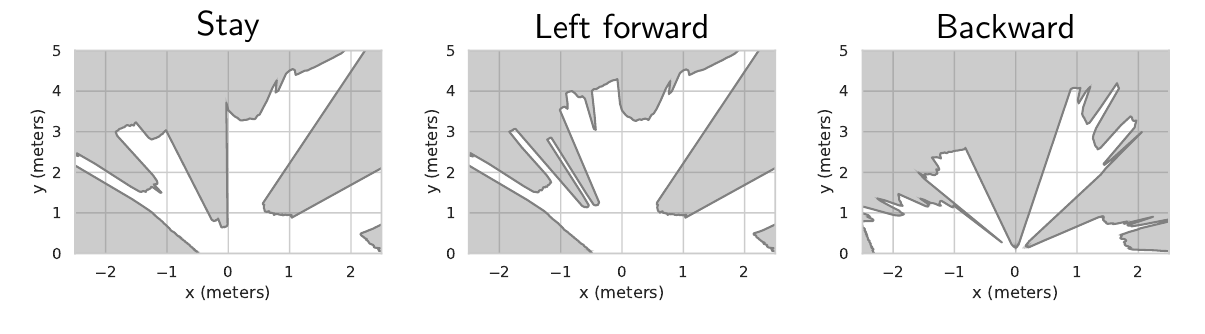}
    \caption{Three training samples of the active-mode following network. The network has to detect feet patterns in laser-rangefinder scans from seven different categories.}
    \label{fig:followmodes}
\end{figure}

Safety level 0 corresponds to standard empirical risk minimization. The other safety levels are enforced via safety-domain training using interval bound propagation (IBP), i.e., certified safety compared to the empirical safety of the vision network. A visualization of a safety-domain from the level 1 specification is shown in Fig. \ref{fig:attacks} on the left.
We collected a total of 2705 training and 570 validation samples, as illustrated in Fig. \ref{fig:followmodes}. the training and validation accuracy is shown in Table \ref{tab:follow_acc}.

For both the vision and follow network, we evaluate each specification in seven standardized scenarios. The scenarios differ in complexity, e.g., operator commands, obstacles, environment, and lighting conditions.
For the vision network, we report the number of misinterpreted gestures, i.e., an enabling or disabling of the controller without the operator's command. Consequently, some types of errors are masked out (for instance, "enabling when the controller is already enabled").

We report a holistic metric for the Follow network if the robot maneuvered correctly for the entire scenario.
The results for the adversarially fine-tuned vision network are shown in Table \ref{tab:adv}. While the network trained with $\varepsilon=1$ performed as well as the model trained by standard ERM, the performance significantly dropped when increasing the adversarial attack budget. 
Given that for human observers, adversarial perturbations with $\varepsilon=2$ are imperceptible, our results indicate that current training methods are unable to enforce non-trivial adversarial robustness on an image classifier in a robotic learning context.

Moreover, the results confirm that the errors occurred uniformly across different scenarios. Therefore, we conclude that imposing adversarial robustness specification causes transient errors confirming our theory.

The results for the safety-domain imposed follow network is reported in Table \ref{tab:follow}. Only the network trained with standard ERM could successfully handle all scenarios. Interestingly, Fig. \ref{fig:attacks} (right image) shows that this network is highly vulnerable to adversarial misclassifications and would output a forward decision if the large object is directly in front of the robot. While the networks with safety-level one and above are immune to such attacks, they perform significantly worse on the seven test-scenarios. With increasing specification level, the performance monotonously decreases until the network trained with the most rigorous safety specification cannot handle any scenario at all.

In contrast to the adversarial experiment of the Vision network, the defects made by the certified networks are conditional errors. In particular, if a network with specification level 1 could not solve a scenario, then a network with 2 and 3 could not either. Moreover, the failure of the level 1 and level 2 networks happened only during forward locomotion, i.e., close to the border of the safety-domains. This observation also supports our claims that errors induced by safety-domain training occur conditionally.

\begin{table}[t]
    \centering
    \caption{Evaluation of the LiDAR follow networks}
    \begin{tabular}{c|c|cccc}
    \toprule
         \# & Scenario & Standard & Safety & Safety & Safety  \\
          & Description &  training & Level 1 & Level 2 & Level 3  \\
          \midrule
         1 & Plain & \cmark & \cmark & \cmark & \xmark \\
         2 & Around boxes & \cmark & \xmark & \xmark & \xmark \\
         3 & Out of corner & \cmark & \cmark & \cmark & \xmark \\
         4 & Through gate & \cmark & \cmark & \xmark & \xmark \\
         5 & Around table & \cmark & \cmark & \cmark & \xmark \\
         6 & Garage parking & \cmark & \cmark & \cmark & \xmark \\
         7 & Narrow hallway & \cmark & \xmark & \xmark & \xmark \\
         \midrule
         Total & &  7/7 & 5/7 & 4/7 & 0/7 \\
         \bottomrule
    \end{tabular}
    \caption*{\footnotesize \textbf{Note:} Evaluation of the LiDAR follow networks with various safety specification enforced on seven standardized test scenarios. Successful navigation of a scenario is marked by a \cmark. Fail indicates unsuccessful tests.}
    \label{tab:follow}
\end{table}

\begin{table}[t]
    \centering
    \caption{Evaluation of the image recognition networks.}
    \begin{tabular}{c|c|ccc}
    \toprule
         \# & Scenario & \multicolumn{3}{c}{Adversarial training radius} \\
         & description & $\varepsilon=0$ & $\varepsilon=1$ & $\varepsilon=2$ \\
         \midrule
         1 & Forward-backward & 1 & - & 5 \\
         2 & With surgical mask & - & 1 & 3 \\
         3 & Against direct sunlight & - & - & 1 \\
         4 & Staying idle & 1 & - & 4 \\
         5 & Summon out of garage & - & 1 & 1 \\
         6 & Artificial lighting (idle) & - & - & 3 \\
         7 & Artificial lighting (follow) & - & - & 1 \\
         \midrule
         Total & & 2 & 2 & 18 \\
         \bottomrule
    \end{tabular}
    \caption*{\footnotesize \textbf{Note:} Evaluation of the image recognition networks trained with and without adversarial training on seven standardized test scenarios. Numbers indicate number of misclassified gestures that triggered a change in operation mode, i.e., errors without an effect are not counted. Dash represent zero misinterpretations.}
    \label{tab:adv}
    \vspace{0mm}
\end{table}

\section{Conclusion}
Adversarial training and its generalization, safety-domain training can, in principle, learn robust and safe deep learning models.
However, in this work, we showed that these methods induce unexplored error profiles in robotic tasks. 
We proposed a framework for characterizing these errors and predicting their presence based on the type of safety specification enforced during the training.

We empirically validated our claims and demonstrated that the type and strictness of the enforced specification govern the real-world performance of learned controllers for robotic environments. 
Our results concluded that adversarial training requires rethinking before being deployed in robot learning \cite{lechner2019designing}.



\section*{ACKNOWLEDGMENT}
M.L. and T.A.H. are supported in part by the Austrian Science Fund (FWF) under grant Z211-N23 (Wittgenstein Award). R.H. and D.R. are partially supported by Boeing. R.G. are partially supported by Horizon-2020 ECSEL Project grant no. 783163 (iDev40). 

\bibliographystyle{IEEEtran} {
    \bibliography{references}  
}

\end{document}